\documentclass[twoside]{article}
\pdfoutput=1 

\usepackage{aistats2020}
\runningtitle{}

\usepackage[numbers]{natbib}  
\usepackage{graphicx}
\usepackage{amsmath}
\usepackage{amssymb}
\usepackage{mathtools}
\usepackage{amsfonts}
\usepackage{comment}
\usepackage{algorithm2e}
\usepackage{algorithmicx}
\usepackage[acronym,nowarn,section,nonumberlist]{glossaries}
\glsdisablehyper{}

\usepackage{xr}
\usepackage{xcolor}
\usepackage{amsthm}

\usepackage[caption = false]{subfig}
\usepackage{multirow}
\usepackage{adjustbox}

\usepackage{"./macro/GrandMacros"}
\usepackage{"./macro/Macro_BIO235"}

\newcommand\inner[2]{\langle #1, #2 \rangle}

\newtheorem{definition}{Definition}
\newtheorem{theorem}{Theorem}
\newtheorem{proposition}{Proposition}

\newtheorem{assumption}{Assumption}

\algblock{ParFor}{EndParFor}
\algnewcommand\algorithmicparfor{\textbf{parfor}}
\algnewcommand\algorithmicpardo{\textbf{do}}
\algnewcommand\algorithmicendparfor{\textbf{end\ parfor}}
\algrenewtext{ParFor}[1]{\algorithmicparfor\ #1\ \algorithmicpardo}
\algrenewtext{EndParFor}{\algorithmicendparfor}

\newacronym{UQ}{UQ}{uncertainty quantification}

\newacronym{RKHS}{RKHS}{reproducing kernel Hilbert space}
\newacronym{RBF}{RBF}{radial basis function}
\newacronym{GP}{GP}{Gaussian process}
\newacronym{CGP}{CGP}{\textit{constrained Gaussian process}}
\newacronym{BNN}{BNN}{Bayesian neural network}
\newacronym{ReLU}{ReLU}{Rectified Linear Unit}

\newacronym{CDF}{CDF}{cumulative distribution function}
\newacronym{PDF}{PDF}{probability density function}
\newacronym{MGF}{MGF}{moment generating function}
\newacronym{MLE}{MLE}{maximum likelihood estimation}
\newacronym{KL}{KL}{Kullback-Leibler}
\newacronym{CvM}{CvM}{Cram\'{e}r von Mises}
\newacronym{iid}{i.i.d.}{independent and identically distributed}

\newacronym{MAP}{MAP}{maximum a posterior}
\newacronym{ELBO}{ELBO}{\emph{evidence lower bound}}
\newacronym{BvM}{BvM}{Bernstein-von Mises}
\newacronym{LAN}{LAN}{Locally Asymptotic Normal}
\newacronym{ARD}{ARD}{automatic relevance determination}

\newacronym{GLM}{GLM}{generalized linear model}
\newacronym{GAM}{GAM}{generalized additive model}
\newacronym{RF}{RF}{random forest}
\newacronym{BART}{BART}{Bayesian additive regression trees}

\newacronym{CRPS}{CRPS}{\textit{continous ranked probability score}}
\newacronym{RMSE}{RMSE}{root mean squared error}
\newacronym{MSE}{MSE}{mean squared error}
\newacronym{CI}{CI}{Coverage Index}
\newacronym{FDR}{FDR}{false discovery rate}

\newacronym{HMC}{HMC}{Hamiltonian Monte Carlo}
\newacronym{MCMC}{MCMC}{Markov Chain Monte Carlo}

\newacronym{BNE}{BNE}{\textit{Bayesian Nonparametric Ensemble}}
\newacronym{CVI}{Calibrated VI}{\textbf{Calibrated Variational Inference}}
\newacronym{LASSO}{LASSO}{least absolute shrinkage and selection Operator}
\newacronym{SCAD}{SCAD}{smoothly clipped absolute deviation}

\begin{document}

%

%

\twocolumn[

\aistatstitle{
Variable Selection with Rigorous Uncertainty Quantification \\
using Deep Bayesian Neural Networks: \\
Posterior Concentration and Bernstein-von Mises Phenomenon
}
\aistatsauthor{Jeremiah Zhe Liu}
\aistatsaddress{
\texttt{zhl112@mail.harvard.edu.}\\
Google Research \& Harvard University*} 
]

\begin{abstract}%
This work develops rigorous theoretical basis for the fact that deep \gls{BNN} is an effective tool for high-dimensional variable selection with rigorous uncertainty  quantification.
We develop new Bayesian non-parametric theorems to show that a properly configured deep \gls{BNN} 
(1) learns the variable importance effectively in high dimensions, 
and its learning rate can sometimes ``break" the curse of dimensionality.
(2) \gls{BNN}'s uncertainty quantification for variable importance is rigorous, in the sense that its $95\%$ credible intervals for variable importance indeed covers the truth $95\%$ of the time (i.e. the \gls{BvM} phenomenon). 
The theoretical results 
suggest a simple variable selection algorithm based on the \gls{BNN}'s credible intervals. Extensive simulation confirms the theoretical findings and shows that the proposed algorithm outperforms existing classic and neural-network-based variable selection methods,   particularly in high dimensions.
\end{abstract}

\vspace{-1em}
\section{Introduction}
\label{sec:intro}
\vspace{-.5em}

The advent of the modern data era has given rise to voluminous, high-dimensional data in which the outcome has complex, nonlinear dependencies on input features. In this nonlinear, high-dimensional regime, a fundamental objective 
is \textit{variable selection}, which refers to the identification of a small subset of features that is relevant in explaining variation in the outcome. 
However, high dimensionality brings two challenges to variable selection. The first is the \textit{curse of dimensionality}, or the exponentially increasing difficulty in learning the variable importance parameters as the dimension of the input features increases.
The second is the impact of \textit{multiple comparisons}, which makes  construction of a high dimensional variable-selection decision rule that maintains an appropriate false discovery rate difficult. For example, consider selecting among 100 variables using a univariate variable-selection procedure that has average precision, defined as 1 - \gls{FDR},  of $0.95$ for selection of a single variable. Then the probability of selecting at least one irrelevant variable out of the $100$ is $1 - 0.95^{100} \approx 0.994$ (assuming independence among decisions), leading to a sub-optimal procedure with precision less than $0.006$ \citep{benjamini_controlling_1995}. 
The multiple comparison problem arises when a multivariate variable-selection decision is made based purely on individual decision rules, ignoring the dependency structure among the decisions across variables. 
This issue arises in a wide variety of application areas, such as genome-wide association studies and portfolio selection  \citep{buhlmann_statistical_2013}, among others.

The objective of this work is to establish deep \glsfirst{BNN} as an effective tool for tackling both of these challenges.
A deep neural network is known to be an effective model for high-dimensional learning problems, illustrating empirical success in image classification and speech recognition applications. 
Bayesian inference in neural networks provides a principled framework for uncertainty quantification that naturally handles the multiple comparison problem \citep{gelman_why_2012}. By sampling from the joint posterior distribution of the variable importance parameters, a deep \gls{BNN}'s 
posterior distribution provides 
a complete picture of the dependency structure among the variable importance estimates for all input variables, 
allowing a variable selection procedure to tailor its decision rule with respect to the correlation structure of the problem. 

Specifically, we propose a simple variable selection method for high-dimensional regression based on credible intervals of a deep \gls{BNN} model. Consistent with the existing nonlinear variable selection literature, we measure the global importance of an input variable $x_p$ using the empirical norm of its gradient function $\psi_p(f)=\|\deriv{x_p}f \|_n^2= \frac{1}{n}\sum_{i=1}^n |\deriv{x_p} f(\bx_i)|^2$, where $f$ is the regression function and $p \in \{1, \dots, P\}$ \citep{white_statistical_2001, rosasco_nonparametric_2013, yang_model-free_2016, he_scalable_2018}. We perform variable selection by first computing the $(1-\alpha)$-level simultaneous credible interval for the joint posterior distribution $\psi(f)=\{\psi_p(f)\}_{p=1}^P$, and make variable-selection decisions by inspecting whether the credible interval includes 0 for a given input. 
Clearly, the validity and effectiveness of this approach hinges critically on a deep \gls{BNN}'s ability to accurately learn and quantify uncertainty about variable importance in high dimensions. Unfortunately, neither property of the a deep \gls{BNN} model is well understood in the literature. 

\textbf{Summary of Contributions}.$\;$ In this work, we establish new Bayesian nonparametric theorems for deep \gls{BNN}s 
to investigate their ability in learning and quantifying uncertainty of variable importance measures derived from the model. We ask two key questions: (1) \textit{learning accuracy}:  does a deep \gls{BNN}'s good performance in prediction (i.e. in learning the true function $f_0$) translate to its ability to learn the variable importance $\psi_p(f_0)$? (2) \textit{uncertainty quantification}: does a deep 
\gls{BNN} properly quantify uncertainty about variable importance, such that a $95\%$ credible interval for variable importance $\psi_p(f)$ covers the ``true" value $\psi_p(f_0)$  $95\%$ of the time?
Our results show that, for \textit{learning accuracy}, a deep Bayesian neural network learns the variable importance at a rate that is at least as fast as that achieved when learning $f_0$ (Theorem \ref{thm:post_convergence}). Furthermore, such rate can sometimes ''break" the curse of dimensionality
, in the sense that the learning rate does not have an exponential dependency on data dimension $P$ 
(Proposition \ref{thm:f_post_converg}). 
For \textit{uncertainty quantification}, we establish a \textit{\glsfirst{BvM} theorem} to show that the posterior distribution of $\psi_p(f)$ converges to a Gaussian distribution, and the $(1-\alpha)$-level credible interval obtained from this distribution covers the true variable importance $\psi_p(f_0)$  $(1-\alpha)\%$ of the time (Theorem \ref{thm:bvm} and \ref{thm:bvm_mvn}). The \gls{BvM} theorems establish a rigorous frequentist interpretation for a deep \gls{BNN}'s simultaneous credible intervals, and are essential in ensuring the validity of the credible-interval-based variable selection methods.
To the authors' knowledge, this is the first semi-parametric \gls{BvM} result for deep neural network models, and therefore the first Bayesian non-parametric study on the deep \gls{BNN}' ability to achieve rigorous uncertainty quantification. 

\textbf{Related Work} The existing variable selection methods for neural networks fall primarily under the frequentist paradigm \citep{anders_model_1999, castellano_variable_2000, guyon_introduction_2003, may_review_2011}. These existing methods include penalized estimation / thresholding of the input weights \citep{feng_sparse_2017, lu_deeppink:_2018, scardapane_group_2017}, greedy elimination based on the perturbed objective function \citep{lecun_optimal_1990, ye_variable_2018}, and re-sampling based hypothesis tests \citep{giordano_input_2014, la_rocca_variable_2005}. For Bayesian inference, the recent work of \cite{liang_bayesian_2018} proposed Spike-and-Slab priors on the input weights and performing variable selection based on the posterior inclusion probabilities for each variable. Rigorous uncertainty quantification based on these approaches can be difficult, due to either the non-identifiability of the neural network weights,  the heavy computation burden of the re-sampling procedure, or the difficulty in developing \gls{BvM} theorems for the neural network model. 

The literature on the theoretical properties of a \gls{BNN} model is relatively sparse. Among the known results, \cite{lee_consistency_2000} established the posterior consistency of a one-layer \gls{BNN} for learning continous or square-integrable functions.  \cite{rockova_posterior_2018} generalized this result to deep architectures and to more general function spaces such as the $\beta$-H\"{o}lder space. Finally, there does not seem to exist a \gls{BvM} result for the deep \gls{BNN} models,  either for the regression function $f$ or a functional of it. Our work addresses this gap by developing a semi-parametric \gls{BvM} theorem  for a multivariate quadratic functional $\psi(f)=\{\|\deriv{x_p}(f)\|_n^2\}_{p=1}^P$ within a deep \gls{BNN} model.

\vspace{-1em}
\section{Background}
\label{sec:background}
\vspace{-1em}
\textbf{Nonparametric Regression} For data $\{y_i, \bx_i\}_{i=1}^n$ where $y \in \real$ and $\bx \in [0, 1]^P$ is a $P \times 1$ vector of covariates, we consider the nonparametric regression setting where $y_i = f^*(\bx_i) + e_i$,  for  $e_i \sim N(0, s^2)$ with known $s$. The data dimension $P$ is allowed to be large but assumed to be $o(1)$. That is,  the dimension does not increase with the sample size $n$.
The data-generation function $f^*$ is an unknown continuous function belonging to certain function class $\Fsc^*$. Recent theoretical work  suggests that the model space of a properly configured deep neural network  $\Fsc(L, K, S, B)$ (defined below) achieves excellent approximation performance for a wide variety of function classes \citep{yarotsky_error_2016, schmidt-hieber_nonparametric_2017, montanelli_new_2017, suzuki_adaptivity_2018, gribonval_approximation_2019}. Therefore in this work, we focus our analysis on the \gls{BNN}'s behavior in learning the optimal $f_0 \in \Fsc(L, K, S, B)$, making an assumption throughout that the \gls{BNN} model is properly configured such that $f_0 \in \Fsc$ is either identical to $f^*$ or is sufficiently close to $f^*$ for practical purposes. 

\textbf{Model Space of a Bayesian Neural Network} Denote $\sigma$ as the \gls{ReLU} activation function. The class of deep \gls{ReLU} neural networks with depth $L$ and width $K$ can be written as
$f(\bx) 
= b_0 + \beta^\top \big[ \sigma \Wsc_L \big(\sigma \Wsc_{L-1} \dots \big( \sigma \Wsc_2 (\sigma \Wsc_1 \bx) \big)\big)\big]$.
Following existing work in deep learning theory, we assume that the hidden weights $\Wsc$ satisfy the sparsity constraint $\Csc^S_0$ and norm constraint $\Csc^B_\infty$ in the sense that:
$\Csc^S_0 = \big\{ 
\Wsc \big| \sum_{l=1}^{L} ||\Wsc_l||_0 \leq S 
\big\}$, $\Csc^B_\infty = \big\{ 
\Wsc \big| \max_{l} ||\Wsc_l||_\infty \leq B, \; B \leq 1 
\big\}$ \citep{schmidt-hieber_nonparametric_2017, suzuki_adaptivity_2018}.
As a result, we denote the class of \gls{ReLU} neural networks with depth $L$, width $K$, sparsity constraint $S$ and norm constraint $B$ as $\Fsc(L, K, S, B)$:
\begin{equation*}
\resizebox{1.1\columnwidth}{!}{%
$\Fsc(L, K, S, B) =
\Big\{
f(\bx) = b_0+\beta^\top \big[ \circ_{l=1}^L (\sigma \Wsc_l) \circ x\big]  \Big| 
\Wsc \in \Csc^S_0, \Wsc \in \Csc^B_\infty 
\Big\},$
}
\label{eq:relu_nn}
\end{equation*}
and for notational simplicity we write $\Fsc(L, K, S, B)$ as $\Fsc$ when it is clear from the context. The Bayesian approach to neural network learning specifies a prior distribution $\Pi(f)$ that assigns probability to every candidate $f$ in the model space $ \Fsc(L, K, S, B)$. The prior distribution $\Pi(f)$ is commonly specified implicitly  through its model weights $\Wsc$,  such that the posterior distribution is $\Pi(f|\{y, \bx\}) \propto \int \Pi(y |\bx, f, \Wsc)\Pi(\Wsc) d\Wsc$.
Common choices for $\Pi(\Wsc)$ include Gaussian \citep{neal_bayesian_1996}, 
Spike and Slab \citep{rockova_posterior_2018}, and Horseshoe priors \citep{ghosh_model_2017, louizos_learning_2018}.

\paragraph{Rate of Posterior Concentration} The quality of a Bayesian learning procedure is commonly measured by the learning rate of its posterior distribution, as defined by the speed at which the posterior distribution $\Pi_n=\Pi(.|\{y_i, \bx_i\}_{i=1}^n)$ shrinks around the truth as $n \rightarrow \infty$. Such speed is usually assessed by the radius of a small ball surrounding $f_0$ that contains the majority of the posterior probability mass. Specifically, we consider the size of a set $A_n = \{f | ||f - f_0||_n \leq M\epsilon_n\}$ such that $\Pi_n(A_n)\rightarrow 1$. Here, the \textit{concentration rate} $\epsilon_n$ describes how fast this small ball $A_n$ concentrates toward $f_0$ as the sample size increases. We state this notion of posterior concentration formally below \citep{ghosal_convergence_2007}:

\begin{definition}[Posterior Concentration]
For $f^*:\real^P \rightarrow \real$ where $P = o(1)$, let $\Fsc(L, K, S, B)$ denote a class of \gls{ReLU} network with depth $L$, width $K$,  sparsity bound $S$ and norm bound $B$. Also denote $f_0$ as the \gls{KL}-projection of $f^*$ to $\Fsc(L, K, S, B)$, and $E_0$ the expectation with respect to true data-generation distribution $P_0 = N(f^*, \sigma^2)$. Then we say the posterior distribution $f$ concentrates around $f_0$ at the rate $\epsilon_n$ in $P^{n}_0$ probability if there exists an $\epsilon_n \rightarrow 0$ such that for any $M_n \rightarrow \infty$:
\begin{align}
E_0\Pi(f: ||f - f_0||^2_n > M_n \epsilon_n | \{y_i, \bx_i\}_{i=1}^n ) \rightarrow 0
\label{eq:f_post_converg}
\end{align}
\vspace{-2em}
\label{def:post_conv}
\end{definition}
\paragraph{``Break" the Curse of Dimensionality} Clearly, a Bayesian learning procedure with good finite-sample performance should have an $\epsilon_n$ that converges quickly to zero. In general, the learning rate $\epsilon_n$ depends on the dimension of the input feature $P$, and the geometry of the ``true" function space $f^* \in \Fsc^*$. Under the typical nonparametric learning scenario where $\Fsc^*$ is the space of $\beta$-H\"{o}lder smooth (i.e., $\beta$-times differentiable) functions, the concentration rate $\epsilon_n$ is found to be $\epsilon_n = O\big( n^{-2\beta/(2\beta + P)} * (log \, n)^{\gamma} \big)$ for some $\gamma > 1$\citep{rockova_posterior_2018}. This exponential dependency of $\epsilon_n$ on the dimensionality $P$ is referred to as the \textbf{\textit{curse of dimensionality}}, which implies that the sample complexity of a neural network explodes exponentially as the data dimension $P$ increases \citep{bach_breaking_2014}. However, recent advances in frequentist learning theory shows that when $f^*$ is sufficiently structured, a neural network model can in fact ``break the curse" by adapting to the underlying structure of the data and achieve a learning rate that has no exponential dependency on $P$ \citep{bach_breaking_2014, suzuki_adaptivity_2018}. 
We show that this result is also possible for Bayesian neural networks when learning $f^*=f_0 \in \Fsc(L,K,S,B)$. That is, this result is possible when the target function lies in the model space of the neural networks.

\textbf{Measure of Variable Importance $\psi_p(f)$}. For a smooth function $f:\real^P \rightarrow \real$, the \textit{local importance} of a variable $x_p$ with respect to the outcome $y=f(\bx)$ at a location $\bx \in \Xsc$ is captured by the magnitude of the \textit{weak}\footnote{The notion of \textit{weak} derivative is a mathematical necessity to ensure $\deriv{x_p} f$ is well-defined, since $f$ involves the \gls{ReLU} function which is piece-wise linear and not differentiable at $0$. However in practice, $\deriv{x_p} f$ can be computed just as a regular derivative function, since it rarely happens that the pre-activation function is exactly $0$.} partial derivative $\big| \deriv{x_p} f(\bx) \big|^2$
\citep{he_scalable_2018, rosasco_nonparametric_2013, wahba_spline_1990, adams_sobolev_2003}. 
Therefore, a natural measure for the \textit{global importance} of a variable $x_p$ is the integrated gradient norm over the entire feature space $\bx \in \Xsc$:
$\Psi_p(f) = \big\| \deriv{x_p} f \big\|_{2}^2 = 
\int_{\bx \in \Xsc} \big| \deriv{x_p} f(\bx) \big|^2 dP(\bx)$.
Given observations $\{\bx_i, y_i\}_{i=1}^n$, $\Psi_p(f)$ is approximated as:
\vspace{-.5em}
\begin{align}
\psi_p(f) =
\big\| \deriv{x_p} f \big\|_n^2 = 
\frac{1}{n}\sum_{i=1}^n \big| \deriv{x_p} f(\bx_i) \big|^2.    
\vspace{-1em}
\end{align}
In practice, $\deriv{x_p} f(\bx)$ can be computed easily using standard automatic differentiation tools \citep{abadi_tensorflow:_2016}.

\section{Learning Variable Importance with Theoretical Guarantee}
\label{sec:theory}
Throughout this theoretical development, we assume the true function $f_0$ has bounded norm $|| f_0 ||_\infty \leq C$, so that the risk minimization problem is well-defined. We also put a weak requirement on the neural network's effective capacity so that the total stochasticity in the neural network prior is manageable:
\begin{assumption}[Model Size] The width of the \gls{ReLU} network model $\Fsc(L, K, S, B)$ does not grow faster than $O(\sqrt{n})$, i.e. $K = o(\sqrt{n})$.
\label{eq:tr_H_bound}
\end{assumption}
This assumption ensures that the posterior estimate for $\psi_p(f)$ is stable in finite samples so that it converges sufficiently quickly 
toward the truth, which is a essential condition for the \gls{BvM} theorem to hold. Assumption \ref{eq:tr_H_bound} is satisfied by most of the popular architectures in practice.
For example, in the ImageNet challenge where there are $1.4 \times 10^7$ images, most of the successful architectures, which include AlexNet, VGGNet, ResNet-152 and Inception-v3, have $K=O(10^3)$ nodes in the output layer
\citep{russakovsky_imagenet_2015, krizhevsky_imagenet_2012, simonyan_very_2014, szegedy_rethinking_2015, he_deep_2016}. Neural networks with fixed architecture also satisfy this requirement, since the growth rate  $o(1)$ for these models is also not faster than $\sqrt{n}$.

\subsection{Rate of Posterior Concentration} 
We first investigate a Bayesian \gls{ReLU} network's ability to accurately learn the variable importance $\Psi_p(f_0)=||\deriv{x_p}(f_0)||^2_2$ in finite samples. We show that, for a \gls{ReLU} network that learns the true function $f_0$ with rate $\epsilon_n$ (in the sense of Definition \ref{def:post_conv}), the \textit{entire} posterior distribution for variable importance $\psi_p(f)$ converges consistently to a point mass at the true $\Psi(f_0)$, at speed not slower than $\epsilon_n$. 
\begin{theorem}[Rate of Posterior Concentration for $\psi_p$]
For $f \in \Fsc(L, K, S, B)$, assuming the posterior distribution $\Pi_n(f)$ concentrates around $f_0$ with rate $\epsilon_n$, the posterior distribution for $\psi_p(f)=||\deriv{x_p}f||^2_n$ contracts toward $\Psi_p(f_0)=||\deriv{x_p}f_0||^2_2$ at a rate not slower than $\epsilon_n$. That is, for any $M_n \rightarrow \infty$
\begin{align*}
    E_0\Pi_n \Big( \sup_{p \in \{1, \dots, P\}} |\psi_p(f) - \Psi_p(f_0)| > M_n \epsilon_n \Big) \rightarrow 0, 
\end{align*}
where $\Pi_n(.) = \Pi(.|\{y_i, \bx_i\}_{i=1}^n)$ denotes the posterior distribution.
\label{thm:post_convergence}
\end{theorem}
A proof for this theorem is in Supplementary Section \ref{sec:post_convergence_proof}. Theorem \ref{thm:post_convergence} confirms two important facts. First, despite the non-identifiablity of the network weights $\Wsc$, a deep \gls{BNN} can reliably recover the variable importance of the true function $\Psi(f_0)$. Second, a deep \gls{BNN} learns the variable importance at least as fast as the rate for learning the prediction function $f_0$. In other words, \textit{good performance in prediction translates to good performance in learning variable importance}. We validate this conclusion in the experiment (Section \ref{sec:exp}), and show that, interestingly, the learning speed for $\Psi_p(f_0)$ is in fact much faster than that for learning $f_0$. Given the empirical success of deep \gls{ReLU} networks in high-dimensional prediction, Theorem \ref{thm:post_convergence} suggests that a \gls{ReLU} network is an effective tool for learning variable importances in high dimension.

\paragraph{Breaking the Curse of Dimensionality} 
Given the statement of Theorem \ref{thm:post_convergence}, it is natural to ask exactly how fast $\epsilon_n$ can go to zero 
for a \gls{BNN} model. To this end, we show that when learning $f_0 \in \Fsc$, a Bayesian \gls{ReLU} network with a standard Gaussian prior can already ``break" the curse of dimensionality and achieve a parametric learning rate of $O(n^{-1/2})$ up to an logarithm factor. 
\begin{proposition}[Posterior Concentration for $f_0 \in \Fsc$]
For the space of \gls{ReLU} network $\Fsc=\Fsc(L, K, S, B)$, assuming
\begin{itemize}
\item the model architecture satisfies:
\begin{align*}
L = O \big( log(N) \big), \quad K = O\big( N \big), \quad S = O\big( N \, log(N) \big), 
\end{align*}
where $N \in \Nbb$ is a function of sample size $n$ such that $log (N) \geq \sqrt{log (n)}$, and 
\item the prior distribution $\Pi(\Wsc)$ is an \gls{iid} product of Gaussian distributions, 
\end{itemize}
then, for $f_0 \in \Fsc$, the posterior distribution $\Pi_n(f)=\Pi(f|\{\bx_i, y_i\}_{i=1}^n)$ contracts toward $f_0$ at a rate of at least $\epsilon_n = O\big( (N/n) * log(N)^3 \big)$. In particular, if $N = o(\sqrt{n})$ (i.e. Assumption \ref{eq:tr_H_bound}), the learning rate is $\epsilon_n = O\big( n^{-1/2} * log(n)^3 \big)$.
\label{thm:f_post_converg}
\end{proposition}
This result appears to be new to the Bayesian neural network literature, and we give a complete full proof in Supplementary Section \ref{sec:lemma_props}. In combination with the result in Theorem \ref{thm:post_convergence}, this result suggests that high-dimensional variable selection in a \gls{BNN} can also ``break" the curse of dimensionality.
We validate this observation empirically in Section \ref{sec:exp}.


\subsection{Uncertainty Quantification}
In this section, we show that the deep  \gls{BNN}'s posterior distribution for variable importance exhibits the \glsfirst{BvM} phenomenon. That is, after proper re-centering, $\Pi_n\big(\psi_p(f)\big)$ converges toward a Gaussian distribution, and the resulting $(1-q)$-level credible intervals achieve the correct coverage for the true variable importance parameters. The \gls{BvM} theorems provide a rigorous theoretical justification for the \gls{BNN}'s ability to quantify its uncertainty about the importance of input variables.

We first explain why the re-centering is necessary. Notice that under noisy observations, $\psi_p(f)=||\deriv{x_p}f||_n^2$ is a quadratic statistic that is strictly positive even when $\psi_p(f_0)=0$. Therefore, the credible interval of un-centered $\psi_p(f)$ will never cover the truth. To this end, it is essential to re-center $\psi_p$ so that it is an unbiased estimate of $\psi(f_0)$: 
\begin{align}
    \psi^c_p(f) = \psi_p(f) - \eta_n. 
\end{align}
Here,  $\eta_n=o_p(\sqrt{n})$ is a de-biasing term that estimates the asymptotically vanishing bias $\psi_p(f_0) - E_0(\psi_p(f))$, whose expression we make explicit in the \gls{BvM} Theorem below. 


\begin{theorem}[\glsfirst{BvM} for $\psi^c_p$]
For $f \in \Fsc(L, W, S, B)$, assume the posterior distribution $\Pi_n(f)$ contracts around $f_0$ at rate $\epsilon_n$. Denote $D_p: f \rightarrow \deriv{x_p}f$ to be the weak differentiation operator, and $H_p = D_p^\top D_p$ the corresponding inner product. For $\bepsilon$ the ``true" noise such that $y = f_0 + \bepsilon$, define
\begin{alignat*}{2}
\hat{\psi}_p &= ||D_p(f_0 + \bepsilon)||_n^2
= 
\psi_p(f_0) + 2\inner{H_pf_0}{\bepsilon}_n + \inner{H_p\bepsilon}{\bepsilon}_n, 
\end{alignat*}
and its centered version as $\hat{\psi}^c_p=\hat{\psi}_p - \hat{\eta}_n$, where $\hat{\eta}_n = tr(\widehat{H}_p)/n$. 
Then the posterior distribution of the centered Bayesian estimator $\psi^c_p(f)=\psi_p(f) - \eta_n$ is asymptotically normal surrounding $\hat{\psi}^c_p$. That is, 
\begin{align*}
\Pi\Big(\sqrt{n}(\psi^c_p(f) - \hat{\psi}^c_p) 
\Big| \{\bx_i, y_i\}_{i=1}^n \Big) \leadsto N(0, 4|| H_pf_0 ||^2_n).
\end{align*}
\label{thm:bvm}
\vspace{-1em}
\end{theorem}
The proof for this result is in Section \ref{sec:bvm_proof}. 
Theorem \ref{thm:bvm} states that the credible intervals from posterior distribution $\Pi_n\big(\psi^c_p(f)\big)$ achieve the correct frequentist coverage in the sense that a $95\%$ credible interval covers the truth $95\%$ of the time. To see why this is the case, notice that a $(1-\alpha)$-level credible set $\hat{B}_n$ under posterior distribution $\Pi_n$ satisfies $\Pi_n\big(\hat{B}_n\big)=1-\alpha$. Also, since $\Pi_n \rightarrow N(0, \sigma^2_{\texttt{BvM}})$, $\hat{B}_n$ also satisfies 
\begin{align}
    \Pi_{N(0,1)}\big( (\hat{B}_n - \hat{\psi}^c_p)/\sigma_{\texttt{BvM}} \big) \rightarrow
    1 - \alpha
    \label{eq:bvm_ci}
\end{align}
in probability for $\sigma_{\texttt{BvM}}^2=4|| H_pf_0 ||^2_n/n$, where $\Pi_{N(0,1)}$ is the standard Gaussian measure. In other words, the set $\hat{B}_n$ can be written in the form of 
$\hat{B}_n = 
\big[
\hat{\psi}^c_p - \rho_\alpha * \sigma_\psi, 
\hat{\psi}^c_p + \rho_\alpha * \sigma_\psi 
\big]$,
which matches the $(1 - \alpha)$-level confidence intervals of an unbiased frequentist estimator $\hat{\psi}_p(f_0)$, which are known to achieve correct coverage for true parameters \citep{van_der_vaart_asymptotic_2000}.


\begin{figure*}[ht]
    \centering
    \subfloat{\includegraphics[width=0.3\textwidth]{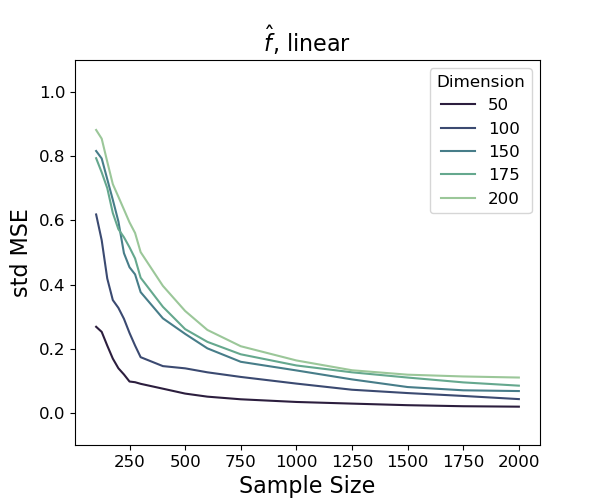}}
    \subfloat{\includegraphics[width=0.3\textwidth]{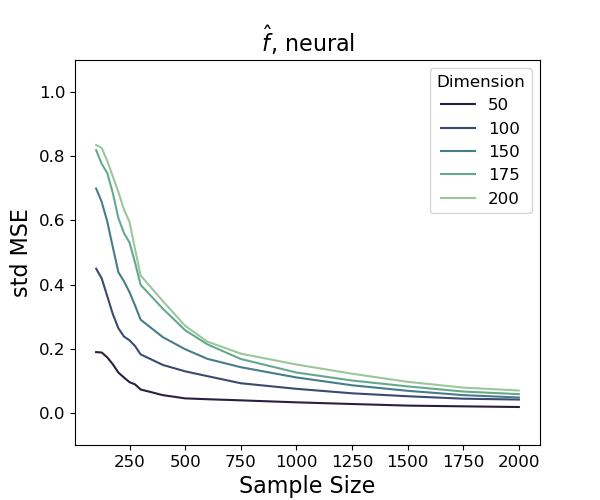}}
    \subfloat{\includegraphics[width=0.3\textwidth]{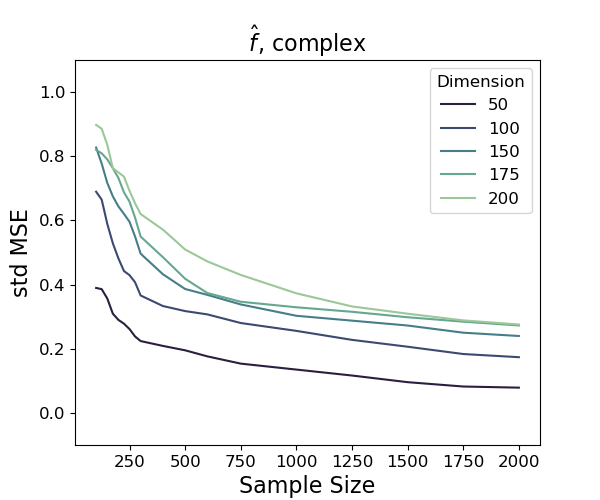}}
    
    \subfloat{\includegraphics[width=0.3\textwidth]{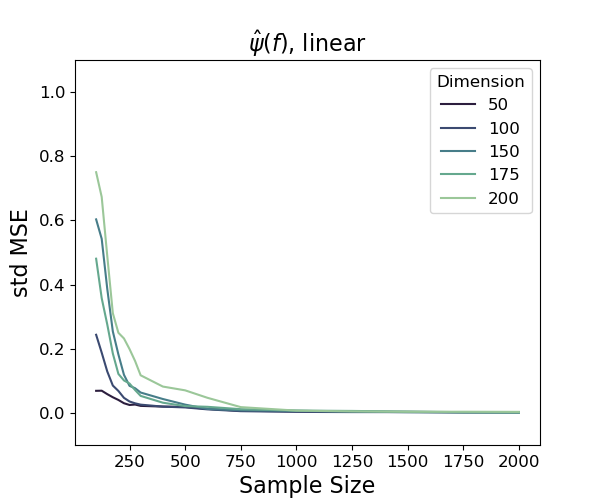}}
    \subfloat{\includegraphics[width=0.3\textwidth]{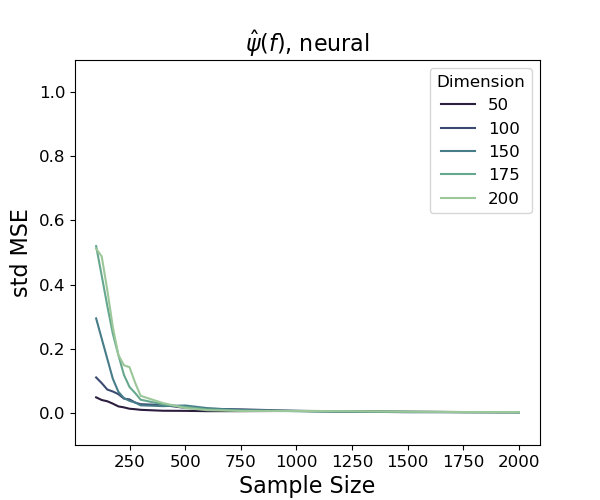}}
    \subfloat{\includegraphics[width=0.3\textwidth]{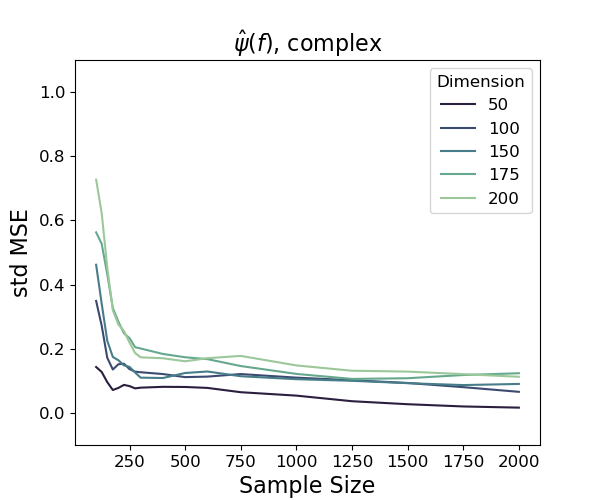}}
\caption{BNN's convergence behavior for learning prediction $f^*$ (first row) and variable importance $\psi(f^*)$ (second row)  under sample sizes $n \in (100, 2000)$ for $P \in (50, 200)$, measured by the standardized MSE (i.e. $1-R^2$). Column 1-3 corresponds to \textbf{linear}, \textbf{neural}, and \textbf{complex}.}

\label{fig:learning}
\end{figure*}

\paragraph{Handling the Issue of Multiple Comparison} Notice that Theorem \ref{thm:bvm} provides justification only for the univariate confidence intervals $\Pi_n(\psi_p^c)$. To handle the issue of \textit{multiple comparisons}, we must take into account the statistical dependencies between all $\{\psi_p^c(f)\}_{p=1}^P$. To this end, in Appendix Section \ref{sec:bvm_mvn}, we extend Theorem \ref{thm:bvm} to the multivariate case to verify that the deep \gls{BNN}'s \textit{simultaneous} credible intervals for all $\{\psi^c_p(f)\}_{p=1}^P$ also have the correct coverage. 
%

\section{Experiment Analysis}
\label{sec:exp}
\subsection{Posterior Concentration and Uncertainty Quantification}
\label{sec:exp_learning}
We first empirically validate the two core theoretic results, 
posterior convergence and Bernstein-von Mises theorem, of this paper. 
In all the experiments described here,  we use the standard \gls{iid} Gaussian priors for model weights, so the model does not have an additional sparse-inducing mechanism beyond \gls{ReLU}. We perform posterior inference using \gls{HMC} with an adaptive step size scheme \citep{andrieu_tutorial_2008}.


\paragraph{Learning Accuracy and Convergence Rate}
We generate data under the Gaussian noise model $y \sim N(f^*, 1)$ for data-generation function $f^*$ with true dimension $P^* = 5$. We vary the 
dimension of the data between $P \in (25, 200)$, and vary sample sizes $n \in (100, 2000)$. 
For the neural network model, we consider a 2-layer, 50-hidden-unit feed-forward architecture (i.e., $L=2$ and $K = 50$) with standard \gls{iid} Gaussian priors $N(0, \sigma^2=0.1)$ for model weights.
We consider three types of data-generating $f^*$: (1) \textbf{linear}:  a simple linear model $f^*(\bx) = \bx^\top \bbeta$; (2)  \textbf{neural}:  a function $f^* \in \Fsc(L, W, S, B)$, and (3) \textbf{complex}:  a complex, non-smooth multivariate function\footnote{$f^*(\bx) = \frac{sin(max(x_1, x_2)) + arctan(x_2)}{1 + x_1 + x_5} + sin(0.5 \, x_3) \big(1 + exp(x_4 - 0.5 \, x_3) \big) + x_3^2 + 2 \, sin(x_4) + 4 \, x_5$, which is non-continuous in terms of $x_1, x_2$ but infinitely differentiable in terms of $x_3, x_4, x_5$
} that is outside the neural network model's approximation space $\Fsc(L, W,S, B)$. This latter data-generating model violates the assumption that $f^* \in \Fsc$ in Proposition \ref{thm:f_post_converg}.
We repeat the simulation 20 times for each setting, and evaluate the neural network's performance in learning $f$ and $\psi_p(f)$ using out-of-sample standardized \gls{MSE}, as follows: 
\vspace{-.5em}
\begin{align*}
&std\_MSE(f, f^*) = \\
&\quad \Big[
\frac{1}{n}\sum_{i=1}^n [f(\bx_i) - f^*(\bx_i)]^2
\Big]\Big/
\Big[
\frac{1}{n}\sum_{i=1}^n[ f^*(\bx_i) - E(f^*(\bx_i))]^2
\Big].
\end{align*}
This is essentially the $1 - R^2$ statistic in regression modeling whose value lies within $(0, 1)$. Use of this statistic allows us to directly compare model performance across different data settings. The $std\_MSE$ for $\psi(f)=\{\psi_p(f)\}_{p=1}^P$ is computed similarly by averaging over all $p \in \{1, \dots, P\}$.

Figure \ref{fig:learning} summarizes the standardized \gls{MSE}s for learning $f^*$ and $\psi(f^*)$, where each column corresponds to a data-generation machanism (\textbf{linear}, \textbf{neural} and \textbf{complex}). The first row summarizes the model's convergence behavior in prediction (learning $f^*$).  We see that the model's learning speed deteriorates as the data dimension $P$ increases. However, this impact of dimensionality appears to be much smaller in the \textbf{linear} and \textbf{neural} scenarios, which both satisfy $f^* \in \Fsc$ (Proposition \ref{thm:f_post_converg}). 
Comparatively, on the second row, the model's learning speed for variable importance are upper bounded by, and in fact \textit{much} faster than, the speed of learning $f^*$. This verifies our conclusion in Theorem \ref{thm:post_convergence} that a model's good behavior in prediction translates to good performance in learning variable importance. We also observe that when the assumption $f^* \in \Fsc$ is violated (e.g. for \textbf{complex} $f^*$ in Column 3), the posterior estimate of $\psi_p(f)$ still converges toward $\psi_p(f_0)$, although at a rate that is much slower and is more sensitive to the dimension $P$ of the data.

\paragraph{Bernstein-von Mises Phenonmenon} We evaluate the \gls{BNN} model's convergence behavior toward the asymptotic posterior $N(0, \sigma_{\texttt{BvM}}^2=4||H_p \, f_0||_n^2)$ using two metrics: (1) the standardized \gls{MSE} for learning the standard deviation $\sigma_{\texttt{BvM}}$, which assesses whether the \textit{spread} of the posterior distribution is correct. (2) The \gls{CvM} statistic as defined as the empirical $L_2$ distance between the standardized posterior sample $\{\psi^c_{std, m}\}_{m=1}^M$ and a Gaussian distribution $\Phi$. This latter statistic,  $CvM(\psi^c_{std}) = 
\frac{1}{M}\sum_{m=1}^M \big[\Fbb(\psi^c_{std, m}) - \Phi(\psi^c_{std, m})\big]^2$,  assesses whether the \textit{shape} of the posterior distribution is sufficiently symmetric and has a Gaussian tail. Notice that since the \gls{CvM} is a quadratic statistic, it roughly follows a mixture of $\chi^2$ distribution even if true variable importance $\psi(f)$ is zero. Therefore,  we compare it against a null distribution of $CvM(\psi^c_{std})$ for which $\psi^c_{std, m}$ is sampled from a Gaussian distribution. 

Figure \ref{fig:uq} in the Appendix summarizes the posterior distribution's convergence behavior in standard deviation (measured by $std\_MSE$, top) and in normality (measured by $CvM$, bottom). The shaded region in the lower figure corresponds to the quantiles of a null CvM distribution. The figure shows that, as the sample size increases, the standardized \gls{MSE} between $sd(\psi^c)$ and $\sigma_{BvM}$ converges toward 0, and the \gls{CvM} statistics enters into the range of the null distribution. The speed of convergence deteriorates as the dimension of the data increases, although not dramatically. These observations indicate that the credible intervals from the variable importance posterior $\Pi_n (\psi^c(f) )$ indeed achieve the correct spread and shape in reasonably large samples, i.e. the Bernstein-von Mises phenomenon 
holds under the neural network model. 

 

\vspace{-1em}
\subsection{Effectiveness in High-dimensional Variable Selection}
\label{sec:exp_selection}

Finally, we study the effectiveness of the proposed variable selection approach (neural variable selection using credible intervals) by comparing it against nine existing methods based on various models (linear-LASSO, random forest, neural network) and decision rules (heuristic thresholding, hypothesis testing, Knockoff). We consider both low- and high-dimension situations ($d \in \{25, 75, 200\}$) and observe how the performance of each variable selection method changes as the sample size grows. 

\begin{table*}[ht]
\caption{Summary of variable selection methods included in the empirical study.}
\begin{adjustbox}{width=0.8\textwidth,center}
\begin{tabular}{|c|c|c|c|}
\hline\hline
\textbf{Model / Metric} & \multicolumn{3}{|c|}{\textbf{Decision Rule}}
\\
\hline\hline
& Thresholding & Hypothesis Test & Knockoff
\\
\hline
Linear Model - LASSO & 
Tibshirani (1996) \citep{tibshirani_regression_1996} & 
Barber and Cand\'{e}s (2015) \citep{barber_controlling_2015} & 
Lockhart et al. (2013)\citep{lockhart_significance_2013}
\\ \hline
Random Forrest - Impurity & 
Breiman (2001)\citep{breiman_random_2001} & 
Cand\'{e}s et al. (2018)\citep{candes_panning_2018} & 
Altmann et al. (2010)\citep{altmann_permutation_2010}
\\ \hline\hline
& Group $L_1$ Thresholding & Spike-and-Slab Probability & Credible Interval
\\ \hline
Neural Network - $\Wsc_1$ & 
Feng and Simon (2018) \citep{feng_sparse_2017} & 
Liang et al. (2018) \citep{liang_bayesian_2018} &
\\ \hline
Neural Network - $\psi^c(f)$ &
 & & 
 (this work)
\\ \hline \hline
\end{tabular}
\end{adjustbox}
\vspace{-1em}
\label{tb:selection_method}
\end{table*}

For the candidate variable selection methods, we notice that a variable selection method usually consists of three components: model, measure of variable importance, and the variable-selection decision rule. To this end, we consider nine methods that span three types of models and three types of decision rules (See Table \ref{tb:selection_method} for a summary). The models we consider are (1) \textbf{LASSO}, the classic linear model $y=\sum_{p=1}^P x_p\beta_p$ with LASSO penalty on regression coefficients $\bbeta$, whose variable importance is measured by the magnitude of $\beta_p$. (2) \textbf{RF}, the random forest model that measures variable importance using \textit{impurity}, i.e., the decrease in regression error due to inclusion of a variable $x_p$ \citep{breiman_random_2001}. (3) \textbf{NNet}, the (deep) neural networks that measure feature importance using either the magnitude of the input weights $\Wsc_1$ or, in our case, the integrated gradient norm $\psi^c(f)$. For \textbf{LASSO} and \textbf{RF}, we consider three types of decision rule: (1) \textbf{Heuristic Thresholding}, which selects a variable by inspecting if the estimate of $\hat{\beta}_p$ is 0 or if the impurity for that variable is greater than $1\%$ of the total impurity summed over  all variables \citep{ye_variable_2018}; (2) \textbf{Knockoff}, a nonparametric inference procedure that controls the \gls{FDR} 
by constructing a data-adaptive threshold for variable importance  \citep{candes_panning_2018}, and (3) \textbf{Hypothesis Test}, which conducts either an asymptotic test on a LASSO-regularized $|\beta_p|$ estimate \citep{lockhart_significance_2013} or permutation-based test based on random forest impurity \citep{altmann_permutation_2010}, For both of these, 
we perform the standard Bonferroni correction. We select the \textbf{LASSO} hyper-parameters $\lambda$ based on 10-fold cross validation, and use 500 regression trees for \textbf{RF}.
For \textbf{NNet}, we also consider three decision rules: the frequentist approach with group-$L_1$ regularization on input weights $\Wsc_1$ \citep{feng_sparse_2017}, a Bayesian approach with spike-and-slab priors on $\Wsc_1$ \citep{liang_bayesian_2018}, and our approach that is based on $95\%$ posterior credible intervals of $\psi_p^c(f)$. Regarding the \textbf{NNet} architecture, we use $L=1, W=5$ for the LASSO- and Spike-and-slab-regularized networks as suggested by the original authors\citep{feng_sparse_2017, liang_bayesian_2018}. We use $L=1, W=50$ for our  approach since it is an architecture that is more common in practice. 

We generate data by sampling the true function from the neural network model $f^* \in \Fsc(L^*=1, W^*=5)$. Notice that this choice puts our method at a disadvantage compared to other \textbf{NNets} methods, since our network width $W=50 > W^*$. We fix the number of data-generating covariates to be $d^*=5$, and perform variable selection on input features $\bX_{n \times P}$ with dimension $P \in \{25, 75, 200\}$ which corresponds to low-, moderate-, and high-dimensional situations.  We vary sample size $n \in (250, 500)$. For each simulation setting $(n, P)$, we repeat the experiment 20 times, and summarize each method's variable selection performance using the $F_1$ score, defined as the  geometric mean of variable selection precision $prec = |\hat{S} \cap S|/|\hat{S}|$ and recall $recl = |\hat{S} \cap S|/|S|$ for $S$ the set of data-generating variables and $\hat{S}$ the set of model-selected variables.

\begin{table*}[ht]
\caption{$F_1$ score for classic and machine-learning based variable selection methods (summarized in Table \ref{tb:selection_method}) under low-dimension (d=25), moderate-dimension (d=75) and high-dimension data (d=200). Boldface indicates the best-performing decision rules in each dimension-model combination.}

\begin{adjustbox}{width=0.75\textwidth,center}
\centering
\begin{tabular}{|c|c|c|c|c|c|c|c|c|}
\hline\hline
 & Model & Rule & n=250 & n=300 & n=350 & n=400 & n=450 & n=500\\
\hline\hline
 \multirow{9}{*}{d=25}
 & \multirow{3}{*}{LASSO} 
   & thres & $0.65 \pm 0.11$ & $0.64 \pm 0.06$ & $0.63 \pm 0.08$ & $0.76 \pm 0.11$ & $0.72 \pm 0.09$ & $0.73 \pm 0.06$ \\
 & & \textbf{knockoff} & $0.99 \pm 0.02$ & $0.99 \pm 0.04$ & $0.94 \pm 0.09$ & $0.98 \pm 0.04$ & $0.99 \pm 0.03$ & $0.99 \pm 0.04$ \\ 
 & & test & $1.00 \pm 0.00$ & $1.00 \pm 0.00$ & $1.00 \pm 0.00$ & $0.89 \pm 0.00$ & $1.00 \pm 0.00$ & $1.00 \pm 0.00$ \\
 \cline{2-9}
 & \multirow{3}{*}{RF} 
   & \textbf{thres} & $1.00 \pm 0.00$ & $1.00 \pm 0.00$ & $1.00 \pm 0.00$ & $1.00 \pm 0.00$ & $1.00 \pm 0.00$ & $1.00 \pm 0.00$ \\
 & & knockoff & $0.62 \pm 0.48$ & $1.00 \pm 0.02$ & $0.96 \pm 0.16$ & $0.90 \pm 0.30$ & $0.94 \pm 0.19$ & $0.99 \pm 0.03$ \\ 
 & & test & $0.91 \pm 0.05$ & $0.98 \pm 0.05$ & $1.00 \pm 0.00$ & $0.98 \pm 0.05$ & $0.98 \pm 0.05$ & $0.98 \pm 0.05$ \\ 
 \cline{2-9}
 & \multirow{3}{*}{NNet} 
   & \textbf{Group $L_1$} & $1.00 \pm 0.00$ & $1.00 \pm 0.00$ & $1.00 \pm 0.00$ & $1.00 \pm 0.00$ & $1.00 \pm 0.00$ & $1.00 \pm 0.00$ \\
 & & SpikeSlab & $0.68 \pm 0.05$ & $0.68 \pm 0.05$ & $0.70 \pm 0.06$ & $0.69 \pm 0.07$ & $0.71 \pm 0.08$ & $0.72 \pm 0.13$ \\  
 & & CI (ours) &  $0.90 \pm 0.04$ & $0.97 \pm 0.05$ & $0.98 \pm 0.04$ & $0.97 \pm 0.05$ & $0.93 \pm 0.06$ & $1.00 \pm 0.00$ \\   
\hline\hline 
\hline\hline
 &  &  & n=250 & n=300 & n=350 & n=400 & n=450 & n=500\\
\hline\hline
 \multirow{9}{*}{d=75}
 & \multirow{3}{*}{LASSO} 
   & thres & $0.32 \pm 0.04$ & $0.31 \pm 0.03$ & $0.31 \pm 0.06$ & $0.46 \pm 0.11$ & $0.56 \pm 0.00$ & $0.53 \pm 0.11$ \\
 & & \textbf{knockoff} & $0.93 \pm 0.14$ & $0.90 \pm 0.14$ & $0.89 \pm 0.15$ & $0.94 \pm 0.08$ & $0.94 \pm 0.11$ & $0.98 \pm 0.04$ \\
 & & test & $0.75 \pm 0.03$ & $0.83 \pm 0.07$ & $0.91 \pm 0.00$ & $0.66 \pm 0.33$ & $0.71 \pm 0.00$ & $0.89 \pm 0.00$ \\
 \cline{2-9}
 & \multirow{3}{*}{RF} 
   & thres & $0.66 \pm 0.10$ & $0.67 \pm 0.06$ & $0.72 \pm 0.10$ & $0.68 \pm 0.06$ & $0.80 \pm 0.04$ & $0.86 \pm 0.04$ \\
 & & knockoff & $0.79 \pm 0.37$ & $0.93 \pm 0.14$ & $0.93 \pm 0.17$ & $0.92 \pm 0.18$ & $0.95 \pm 0.09$ & $0.98 \pm 0.05$ \\ 
 & & \textbf{test} & $0.89 \pm 0.12$ & $0.93 \pm 0.07$ & $0.86 \pm 0.04$ & $0.88 \pm 0.07$ & $0.90 \pm 0.09$ & $0.95 \pm 0.05$ \\ 
 \cline{2-9}
 & \multirow{3}{*}{NNet} 
   & Group $L_1$ & $0.77 \pm 0.00$ & $0.67 \pm 0.27$ & $0.68 \pm 0.23$ & $0.77 \pm 0.00$ & $0.77 \pm 0.00$ & $0.77 \pm 0.00$ \\  
 & & SpikeSlab & $0.63 \pm 0.09$ & $0.66 \pm 0.06$ & $0.65 \pm 0.08$ & $0.65 \pm 0.06$ & $0.67 \pm 0.07$ & $0.68 \pm 0.10$ \\  
 & & \textbf{CI (ours)} & $0.98 \pm 0.04$ & $0.97 \pm 0.04$ & $0.91 \pm 0.07$ & $0.97 \pm 0.04$ & $0.98 \pm 0.05$ & $1.00 \pm 0.00$ 
\\   
\hline\hline 
\hline\hline
 &  &  & n=250 & n=300 & n=350 & n=400 & n=450 & n=500\\
\hline\hline
 \multirow{9}{*}{d=200}
 & \multirow{3}{*}{LASSO} 
   & thres & $0.29 \pm 0.05$ & $0.32 \pm 0.01$ & $0.28 \pm 0.05$ & $0.38 \pm 0.10$ & $0.42 \pm 0.08$ & $0.35 \pm 0.06$ \\
 & & \textbf{knockoff} & $0.31 \pm 0.42$ & $0.68 \pm 0.38$ & $0.88 \pm 0.21$ & $0.89 \pm 0.11$ & $0.90 \pm 0.09$ & $0.87 \pm 0.18$ \\ 
 & & test & $0.21 \pm 0.04$ & $0.25 \pm 0.03$ & $0.04 \pm 0.00$ & $0.49 \pm 0.02$ & $0.27 \pm 0.13$ & $0.61 \pm 0.04$ \\ 
 \cline{2-9}
 & \multirow{3}{*}{RF} 
   & thres & $0.37 \pm 0.02$ & $0.42 \pm 0.01$ & $0.43 \pm 0.06$ & $0.52 \pm 0.02$ & $0.54 \pm 0.05$ & $0.59 \pm 0.05$ \\
 & & knockoff & $0.12 \pm 0.25$ & $0.29 \pm 0.39$ & $0.38 \pm 0.42$ & $0.70 \pm 0.42$ & $0.80 \pm 0.39$ & $0.44 \pm 0.49$ \\
 & & \textbf{test} & $0.79 \pm 0.10$ & $0.81 \pm 0.13$ & $0.79 \pm 0.07$ & $0.87 \pm 0.11$ & $0.83 \pm 0.09$ & $0.70 \pm 0.08$ \\ 
 \cline{2-9}
 & \multirow{3}{*}{NNet} 
   & Group $L_1$ & $0.67 \pm 0.00$ & $0.67 \pm 0.00$ & $0.67 \pm 0.00$ & $0.67 \pm 0.00$ & $0.67 \pm 0.00$ & $0.67 \pm 0.00$ \\
 & & SpikeSlab & $0.45 \pm 0.26$ & $0.53 \pm 0.17$ & $0.57 \pm 0.14$ & $0.60 \pm 0.14$ & $0.57 \pm 0.12$ & $0.57 \pm 0.11$ \\
 & & \textbf{CI (ours)} & $0.84 \pm 0.10$ & $0.76 \pm 0.08$ & $0.84 \pm 0.08$ & $0.93 \pm 0.07$ & $0.98 \pm 0.04$ & $0.92 \pm 0.08$ \\   
\hline\hline 
\end{tabular}
\end{adjustbox}
\vspace{-1em}
\label{tb:selection_f1}
\end{table*}

Table \ref{tb:selection_f1} summarizes the performance as quantified by the $F1$ score of the variable-selection methods in low-, medium- and high-dimension situations. In general, we observe that across all methods, \textbf{LASSO-knockoff}, \textbf{RF-test} and our proposed \textbf{NNet-CI}  tend to have good performance, with \textbf{NNet-CI} being more effective in higher dimensions (d=200). 

Our central conclusion is that \textbf{a powerful model alone is not sufficient to guarantee effective variable selection}. A good measure of variable importance, in terms of an unbiased and low-variance estimator of the true variable importance, and also a rigorous decision rule that has performance guarantee in terms of control over \gls{FDR} or Type-I error are equally important. For example, although based on a neural network that closely matches the truth, \textbf{NNet-Group $L_1$} and \textbf{NNet-SpikeSlab}  measures variable importance using the input weight $\widehat{\Wsc}_1$, which is an unstable estimate of variable importance due to over-parametrization and/or non-identifiablity.
As a result, the performance of these two models are worse than the linear-model based \textbf{LASSO-knockoff}.
Comparing between the decision rules, the heuristic thresholding rules (\textbf{LASSO-thres} and \textbf{RF-thres}) are mostly not optimized for variable selection performance. As a result, they tend to be susceptible to the multiple comparison problem and their performance deteriorates quickly as the dimension increases. The Knockoff-based methods (\textbf{LASSO-knockoff} and \textbf{RF-knockoff}) are nonparametric procedures that are robust to model misspecification but tend to have weak power when the model variance is high. As a result, the Knockoff approach produced good results for the low-variance linear-LASSO model, but comparatively worse result for the more flexible but high-variance random forest model. Finally, the hypothesis tests / credible intervals are model-based procedures whose performance depends on the quality of the model. Hypothesis tests are expected to be more powerful when the model yields an unbiased and low-variance estimate of $f^*$ (i.e. \textbf{RF-test} and \textbf{NNet-CI}), but has no performance guarantee when the model is misspecified (i.e. \textbf{LASSO}). In summary, we find that the \textbf{NNet-CI} method combines a powerful model that is effective in high dimension with a good variable-importance measure that has fast rate of convergence and also a credible-based selection rule that has a rigorous statistical guarantee. As a result, even without any sparse-inducing model regularization, \textbf{NNet-CI} out-performed its \textbf{NNet}-based peers, and is more powerful than other \textbf{LASSO}- or \textbf{RF}-based approaches in high dimensions.

\vspace{-1em}
\section{Discussion and Future Directions}
\label{sec:discussion}
\vspace{-.5em}

In this work, we investigate the theoretical basis underlying the deep \gls{BNN}'s ability to achieve rigorous uncertainty quantification in variable selection. 
Using the square integrated gradient $\psi_p(f)=||\deriv{x_p}f||_n^2$ as the measure of variable importance, we established two new Bayesian nonparametric results on the \gls{BNN}'s ability to learn and quantify uncertainty about variable importance. Our results suggest that the neural network can learn variable importance effectively in high dimensions (Theorem \ref{thm:post_convergence}), in a speed that in some cases ``breaks" the curse of dimensionality (Proposition \ref{thm:f_post_converg}).  Moreover, it can generate rigorous and calibrated uncertainty estimates in the sense that its $(1-q)$-level credible intervals for variable importance cover the true parameter $(1-q)\%$ of the time (Theorem \ref{thm:bvm}-\ref{thm:bvm_mvn}). The simulation experiments confirmed these theoretical findings, and revealed the interesting fact that \gls{BNN} can learn variable importance $\psi_p(f)$ at a rate much faster than learning predictions for $f^*$ (Figure \ref{fig:learning}). The comparative study illustrates the effectiveness of the proposed approach for the purpose of variable selection in high dimensions, which is a scenario where the existing methods experience difficulties due to model misspecification, the curse of dimensionality, or the issue of multiple comparisons. 

Consistent with classic Bayesian nonparametric and deep learning literature
\citep{castillo_bernsteinvon_2015, rockova_theory_2018, barron_universal_1993, barron_approximation_2018}, the theoretical  results developed in this work assume a well-specified scenario where $f^* \in \Fsc$ and the noise distribution is known. Furthermore, computing exact credible intervals 
under a \gls{BNN} model requires the use of 
MCMC
procedures, which can be infeasible for large datasets. 
Therefore two important future directions of this work are to investigate the \gls{BNN}'s ability to learn variable importance under model misspecification,
and to identify posterior inference methods (e.g., particle filter \citep{dai_provable_2016} or structured variational inference
\citep{pradier_projected_2018}) that are scalable to large datasets and can also achieve rigorous uncertainty quantification.

\clearpage
\section*{Acknowledgements}

I would like to thank Dr. Rajarshi Mukherjee and my advisor Dr. Brent Coull at Harvard Biostatistics for the helpful discussions and generous support. 
This publication was made possible by USEPA grant RD-83587201  Its contents are solely the responsibility of the grantee and does not necessarily represent the official views of the USEPA.  Further, USEPA does not endorse the purchase of any commercial products or services mentioned in the publication.


\bibliographystyle{unsrt}

\appendix

\section{Statement of Multivariate \gls{BvM} Theorem}
\label{sec:bvm_mvn}

\begin{theorem}[Multivariate \glsfirst{BvM} for $\psi^c$]
For $f \in \Fsc(L, W, S, B)$, assuming the posterior distribution $\Pi_n(f)$ contracts around $f_0$ at rate $\epsilon_n$.
Denote $\hat{\psi}^c=[\hat{\psi}^c_{1}, \dots, \hat{\psi}^c_P]$ for $\hat{\psi}^c_p$ as defined in Theorem \ref{thm:bvm}. Also recall that $P=o(1)$, i.e. the data dimension does not grow with sample size.

Then $\hat{\psi}^c$ is an unbiased estimator of $\psi(f_0)=[\psi_1(f_0), \dots, \psi_P(f_0)]$, and the posterior distribution for $\psi^c(f)$ asymptotically converge toward a multivariate normal distribution surrounding $\hat{\psi}^c$, i.e.
\begin{align*}
\Pi\Big(\sqrt{n}(\psi^c(f) - \hat{\psi}^c) 
\Big| \{\bx_i, y_i\}_{i=1}^n \Big) \leadsto MVN(0, V_0),
\end{align*}
where $V_0$ is a $P \times P$ matrix such that $(V_0)_{p_1, p_2} =  4\inner{H_{p_1}f_0}{H_{p_2}f_0}_n$.
\label{thm:bvm_mvn}
\end{theorem}
Proof is in Supplementary Section \ref{sec:bvm_mvn_proof}.

\section{Table and Figures}
\begin{figure}[ht]
    \centering
    \subfloat{\includegraphics[width=0.95\columnwidth]{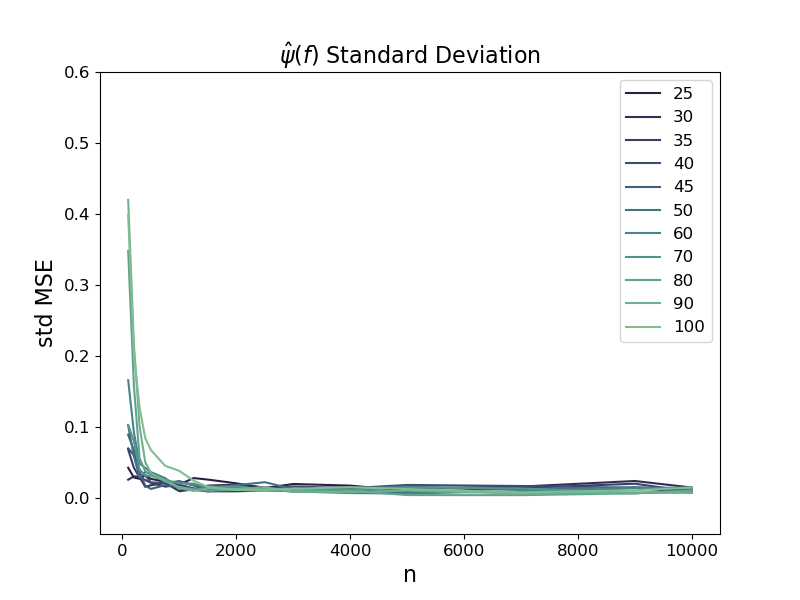}}

    \subfloat{\includegraphics[width=0.95\columnwidth]{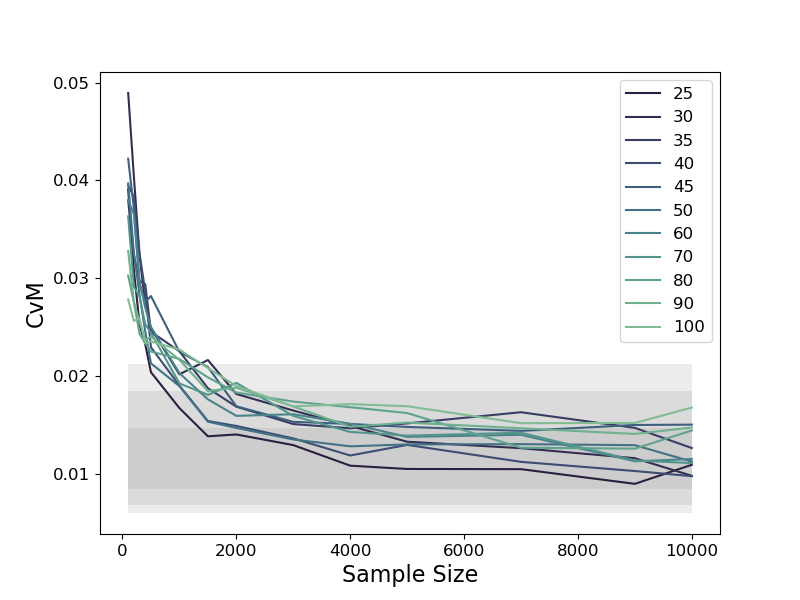}}
\caption{The variable importance posterior's convergence behavior toward the asymptotic standard deviation (left, measured by standardized \gls{MSE}) and toward normality (right, measured by the \gls{CvM} distance from a Gaussian distribution) under sample size $n \in (100, 10000)$ and $P \in (25, 100)$. Shaded region in the right figure indicates the $\{5\%, 10\%, 25\%, 75\%, 90\%, 95\%\}$ quantiles of the null CvM distribution.}
\label{fig:uq}
\end{figure}

%
%
\end{document}